\documentclass[11pt,a4paper]{article}
\usepackage[hyperref]{acl2021}
\usepackage{times}
\usepackage{latexsym}
\usepackage{multirow}
\usepackage{graphicx}
\usepackage{booktabs}
\usepackage{paralist}
\usepackage{amsmath}
\usepackage{subcaption}
\usepackage[normalem]{ulem}
\usepackage{tabulary}

\usepackage{microtype}

\aclfinalcopy 
\def\aclpaperid{1737} 

\title{Robustness Testing of Language Understanding in Task-Oriented Dialog}

\author{Jiexi Liu$^{1*}$, Ryuichi Takanobu$^{1*}$, Jiaxin Wen$^1$, Dazhen Wan$^1$, \\
{\bf Hongguang Li$^2$, Weiran Nie$^2$, Cheng Li$^2$, Wei Peng$^2$, Minlie Huang$^{1\dagger}$} \\
 $^1$CoAI Group, DCST, IAI, BNRIST, Tsinghua University, Beijing, China\\
 $^2$Huawei Technologies, Shenzhen, China \\
  {\small \tt \{liujiexi19, gxly19\}@mails.tsinghua.edu.cn, aihuang@tsinghua.edu.cn} \\
}

\date{}

\begin{document}
\maketitle
\begin{abstract}
  Most language understanding models in task-oriented dialog systems are trained on a small amount of annotated training data, and evaluated in a small set from the same distribution. However, these models can lead to system failure or undesirable output when being exposed to natural language perturbation or variation in practice. In this paper, we conduct comprehensive evaluation and analysis with respect to the robustness of natural language understanding models, and introduce three important aspects related to language understanding in real-world dialog systems, namely, \textit{language variety}, \textit{speech characteristics}, and \textit{noise perturbation}. We propose a model-agnostic toolkit LAUG to approximate natural language perturbations for testing the robustness issues in task-oriented dialog. Four data augmentation approaches covering the three aspects are assembled in LAUG, which reveals critical robustness issues in state-of-the-art models. The augmented dataset through LAUG can be used to facilitate future research on the robustness testing of language understanding in task-oriented dialog.
\end{abstract}
\renewcommand{\thefootnote}{\fnsymbol{footnote}}
\footnotetext[1]{Equal contribution.}
\footnotetext[2]{Corresponding author.}
\renewcommand{\thefootnote}{\arabic{footnote}}

\section{Introduction}
\label{sec:intro}
Recently task-oriented dialog systems have been attracting more and more research efforts \cite{gao2019neural,zhang2020recent}, 
where understanding user utterances is a critical precursor to the success of such dialog systems. While modern neural networks have achieved state-of-the-art results on language understanding (LU) \cite{wang2018bi,zhao2018improving,goo2018slot,liu2019cm,shah2019robust}, their robustness to changes in the input distribution is still one of the biggest challenges in practical use. 


Real dialogs between human participants involve language phenomena that do not contribute so much to the intent of communication. As shown in Fig. \ref{fig:intro}, user expressions can be of high lexical and syntactic diversity when a system is deployed to users; typed texts may differ significantly from those recognized from voice speech; interaction environments may be full of chaos and even users themselves may introduce irrelevant noises such that the system can hardly get clean user input.

Unfortunately, neural LU models are vulnerable to these natural perturbations that are legitimate inputs but not observed in training data. 
For example, \citet{bickmore2018patient} found that popular conversational assistants frequently failed to understand real health-related scenarios and were unable to deliver adequate responses on time. 
Although many studies have discussed the LU robustness \cite{ray2018robust,zhu2018robust,iyyer2018adversarial,yoo2019data,ren2019generating,jin2020bert,he2020learning}, there is a lack of systematic studies for real-life robustness issues and corresponding benchmarks for evaluating task-oriented dialog systems.

In order to study the real-world robustness issues, we define the LU robustness from three aspects: \textit{language variety}, \textit{speech characteristics} and \textit{noise perturbation}. While collecting dialogs from deployed systems could obtain realistic data distribution, it is quite costly and not scalable since a large number of conversational interactions with real users are required.
Therefore, we propose an automatic method LAUG for \textbf{L}anguage understanding \textbf{AUG}mentation in this paper to approximate the natural perturbations to existing data. LAUG is a black-box testing toolkit on LU robustness composed of four data augmentation methods, including word perturbation, text paraphrasing, speech recognition, and speech disfluency. 

We instantiate LAUG on two dialog corpora Frames \cite{el2017frames} and MultiWOZ \cite{budzianowski2018multiwoz} to demonstrate the toolkit's effectiveness. Quality evaluation by annotators indicates that the utterances augmented by LAUG are reasonable and appropriate with regards to each augmentation approach's target. A number of LU models with different categories and training paradigms are tested as base models with in-depth analysis. Experiments indicate a sharp performance decline in most baselines in terms of each robustness aspect. Real user evaluation further verifies that LAUG well reflects real-world robustness issues. Since our toolkit is model-agnostic and does not require model parameters or gradients, the augmented data can be easily obtained for both training and testing to build a robust dialog system. 

Our contributions can be summarized as follows: 
    (1) We classify the LU robustness systematically into three aspects that occur in real-world dialog, including language variety, speech characteristics and noise perturbation;
    (2) We propose a general and model-agnostic toolkit, \textit{LAUG}, which is an integration of four data augmentation methods on LU that covers the three aspects.
    (3) We conduct an in-depth analysis of LU robustness on two dialog corpora with a variety of baselines and standardized evaluation measures. 
    (4) Quality and user evaluation results demonstrate that the augmented data are representative of real-world noisy data, therefore can be used for future research to test the LU robustness in task-oriented dialog\footnote{The data, toolkit, and codes are available at \url{https://github.com/thu-coai/LAUG}, and will be merged into \url{https://github.com/thu-coai/ConvLab-2} \cite{zhu2020convlab}.}.

\section{Robustness Type}

\begin{figure}[!thb]
    \centering
    \begin{subfigure}{0.9\linewidth}\centering
    \includegraphics[width=\linewidth]{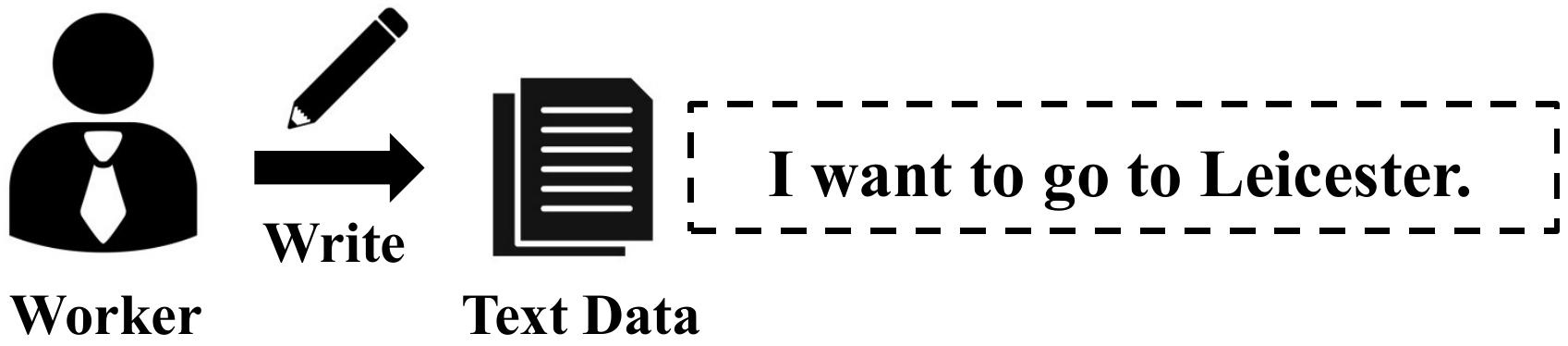}
    \caption{Dataset construction}
    \end{subfigure}
    \begin{subfigure}{\linewidth}\centering
    \includegraphics[width=\linewidth]{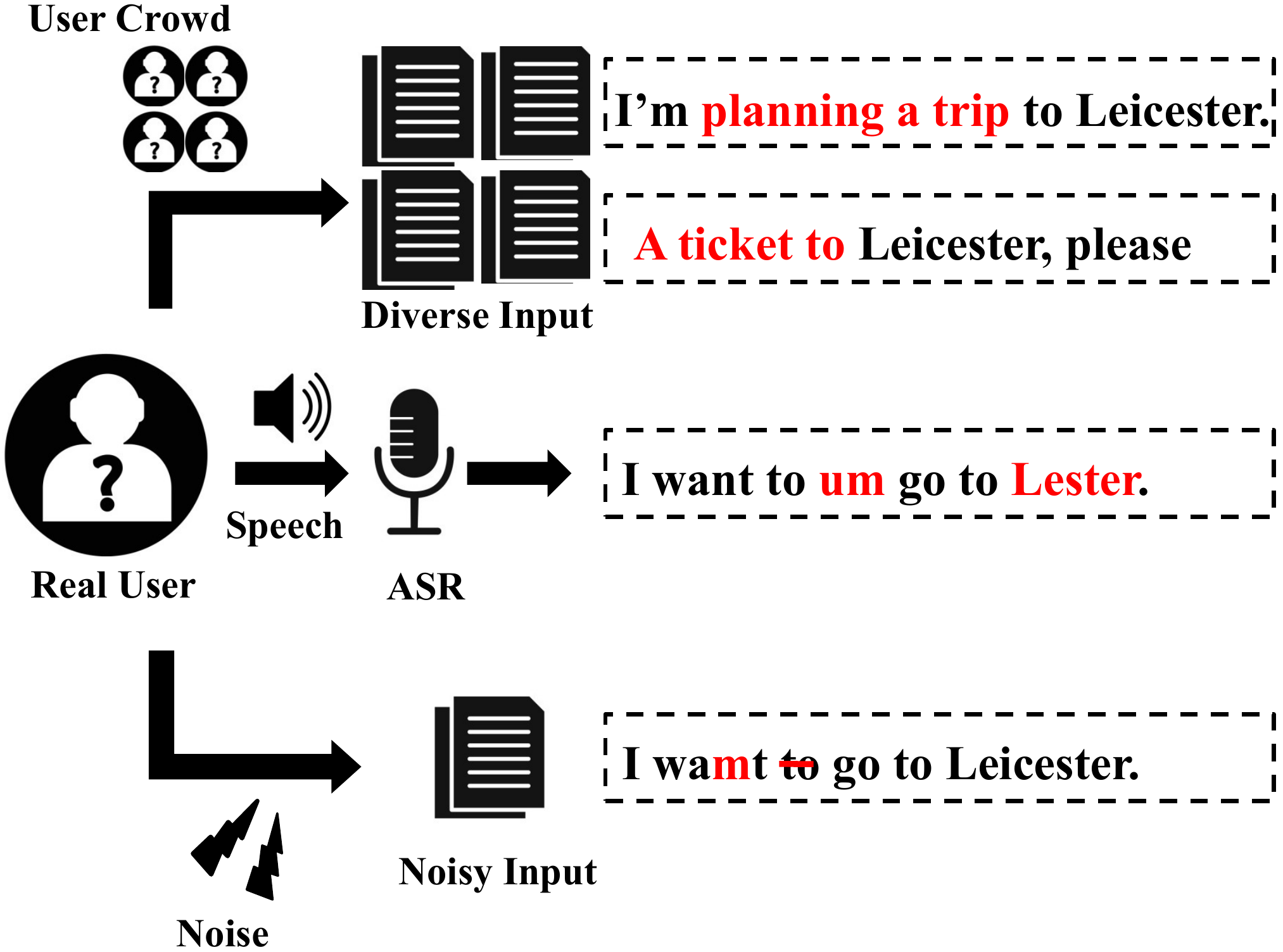}
    \caption{Real-world application}\label{fig:type}
    \end{subfigure}
    \caption{Difference between dialogs collected for training and those for real-world applications.}
    \label{fig:intro}
\end{figure}

We summarize several common interleaved challenges in language understanding from three aspects, as shown in Fig. \ref{fig:type}:

\paragraph{Language Variety}
A modern dialog system in a text form has to interact with a large variety of real users. The user utterances can be characterized by a series of linguistic phenomena with a long tail of variations in terms of spelling, vocabulary, lexical/syntactic/pragmatic choice \cite{ray2018robust,jin2020bert,he2020learning,zhao2019data,ganhotra2020effects}. 

\paragraph{Speech Characteristics}
The dialog system can take voice input or typed text, but these two differ in many ways. For example, written language tends to be more complex and intricate with longer sentences and many subordinate clauses, whereas spoken language can contain repetitions, incomplete sentences, self-corrections and interruptions \cite{wang2020data,park2019specaugment,wang2020multi,honal2003correction,zhu2018robust}. 

\paragraph{Noise Perturbation}
Most dialog systems are trained only on noise-free interactions.
However, there are various noises in the real world, including background noise, channel noise, misspelling, and grammar mistakes \cite{xu2014targeted,li2020textat,yoo2019data,henderson2012discriminative,ren2019generating}. 

\section{LAUG: Language Understanding Augmentation}
This section introduces commonly observed out-of-distribution data in real-world dialog into existing corpora. We approximate natural perturbations in an automatic way instead of collecting real data by asking users to converse with a dialog system.

To achieve our goals, we propose a toolkit \textit{LAUG}, for black-box evaluation of LU robustness. It is an ensemble of four data augmentation approaches, including Word Perturbation (WP), Text Paraphrasing (TP), Speech Recognition (SR), and Speech Disfluency (SD). Noting that LAUG is model-agnostic and can be applied to any LU dataset theoretically. 
Each augmentation approach tests one or two proposed aspects of robustness as Table \ref{tab:check} shows. The intrinsic evaluation of the chosen approaches will be given in Sec. \ref{sec:setup}.

\begin{table}[h!tb]
    \centering
    \small
    \begin{tabular}{c|ccc}
    \hline
        Capacity & LV & SC & NP \\
    \hline
        Word Perturbation (WP) & $\surd$ & & $\surd$ \\
        Text Paraphrasing (TP) & $\surd$ & &\\
        Speech Recognition (SR) &  & $\surd$ & $\surd$ \\
        Speech Disfluency (SD) &  & $\surd$ & \\
    \hline
    \end{tabular}
    \caption{The capacity that each augmentation method evaluates, including Language Variety (LV), Speech Characteristics (SC) and Noise Perturbation (NP).}
    \label{tab:check}
\end{table}

\paragraph{Task Formulation}
Given the dialog context $X_t = \{x_{2t-m}, \dots, x_{2t-1}, x_{2t}\}$ at dialog turn $t$, where each $x$ is an utterance and $m$ is the size of sliding window that controls the length of utilizing dialog history, the model should recognize $y_t$, the dialog act (DA) of $x_{2t}$. Empirically, we set $m = 2$ in the experiment. 
Let $\mathcal{U}, \mathcal{S}$ denote the set of user/system utterances, respectively. Then, we have $x_{2t-2i} \in \mathcal{U}$ and $ x_{2t-2i-1} \in \mathcal{S}$. The task of this paper is to examine different LU models whether they can predict $y_t$ correctly given a perturbed input $\tilde{X}_t$. The perturbation is only performed on user utterances.

\paragraph{Word Perturbation}

Inspired by EDA (\textit{Easy Data Augmentation}) \cite{wei2019eda}, we propose its semantically conditioned version, SC-EDA, which considers task-specific augmentation operations in LU. SC-EDA injects word-level perturbation into each utterance $x'$ and updates its corresponding semantic label $y'$.



\begin{table}[h!tb]
    \small
    \begin{tabular}{c | p{5.8cm}}
    \hline
        Original & I want to go to Cambridge . \\ 
        DA & attraction \{ inform (dest = Cambridge) \} \\
    \hline
        Syno. & I \textcolor{red}{wishing} to go to Cambridge . \\ 
        Insert & I \textcolor{red}{need} want to go to Cambridge . \\ 
        Swap & I \textcolor{red}{to want} go to Cambridge . \\ 
        Delete & I want \sout{\textcolor{red}{to}} go to Cambridge . \\ 
    \hline
        SVR & I want to go to \textcolor{red}{Liverpool} . \\
        DA & attraction \{ inform (dest = \textcolor{red}{Liverpool}) \} \\
    \hline
    \end{tabular}
    \caption{An SC-EDA example. Syno., Insert, Swap and Delete are four operations described in EDA, of which the dialog act is identical to the original one. SVR denotes \textit{slot value replacement}.}
    \label{tab:todeda}
\end{table}

Table \ref{tab:todeda} shows an example of SC-EDA. Original EDA randomly performs one of the four operations, including \textit{synonym replacement}, \textit{random insertion}, \textit{random swap} and \textit{random deletion}\footnote{See the EDA paper for details of each operation.}. Noting that, to keep the label unchanged, words related to slot values of dialog acts are not modified in these four operations.
Additionally, we design \textit{slot value replacement}, which changes the utterance and label at the same time to test model's generalization to \textbf{unseen entities}. 
Some randomly picked slot values are replaced by unseen values with the same slot name in the database or crawled from web sources. For example in Table \ref{tab:todeda}, ``Cambridge'' is replaced by ``Liverpool'', where both belong to the same slot name ``dest'' (destination).



\textit{Synonym replacement} and \textit{slot value replacement} aim at increasing the language variety, while \textit{random word insertion/deletion/swap} test the robustness of noise perturbation. From another perspective, four operations from EDA perform an Invariance test, while \textit{slot value replacement} conducts a Directional Expectation test according to CheckList \cite{ribeiro2020beyond}.

\paragraph{Text Paraphrasing}
The target of text paraphrasing is to generate a new utterance $x' \neq x$ while maintaining its dialog act unchanged, i.e. $y' = y$. 
We applied SC-GPT \cite{peng2020few}, a fine-tuned language model conditioned on the dialog acts, to paraphrase the sentences as data augmentation. Specifically, it characterizes the conditional probability $p_\theta(x|y) = \prod_{k=1}^{K} p_\theta (x_k | x_{<k}, y),$
where $x_{<k}$ denotes all the tokens before the $k$-th position. The model parameters $\theta$ are trained by maximizing the log-likelihood of $p_\theta$.

\begin{table}[thb]
    \centering
    \small
    \begin{tabular}{@{ }c@{ }|@{ }p{6.8cm}@{ }}
    \hline
        DA & train \textcolor{red}{*} \{ inform ( dest = Cambridge ; arrive = 20:45 ) \} \\
        Text & Hi, I'm looking for a train that is going to Cambridge and arriving there by 20:45, is there anything like that? \\
    \hline
        DA & train \{ inform ( dest = Cambridge ; arrive = 20:45 ) \} \\
        Text & Yes, to Cambridge, and I would like to arrive by 20:45. \\
    \hline
    \end{tabular}
    \caption{A pair of examples that consider contextual resolution or not. In the second example, the user omits to claim that he wants a train in the second utterance since he has mentioned this before.}
    \label{tab:paraphrase}
\end{table}

We observe that co-reference and ellipsis frequently occurs in user utterances. Therefore, we propose different encoding strategies during paraphrasing to further evaluate each model's capacity for \textbf{context resolution}. In particular, if the user mentions a certain domain \textit{for the first time} in a dialog, we will insert a ``*'' mark into the sequential dialog act $y'$ to indicate that the user tends to express without co-references or ellipsis, as shown in Table \ref{tab:paraphrase}. Then SC-GPT is finetuned on the processed data so that it can be aware of dialog context when generating paraphrases. As a result, we find that the average token length of generated utterances with/without ``*'' is 15.96/12.67 respectively after SC-GPT's finetuning on MultiWOZ.

It should be noted that slot values of an utterance can be paraphrased by models, resulting in a different semantic meaning $y'$. To prevent generating irrelevant sentences, we apply automatic value detection in paraphrases with original slot values by fuzzy matching\footnote{\url{https://pypi.org/project/fuzzywuzzy/}}
, and replace the detected values in bad paraphrases with original values. In addition, we filter out paraphrases that have missing or redundant information compared to the original utterance.

\paragraph{Speech Recognition} \label{sec:SR}
We simulate the speech recognition (SR) process with a TTS-ASR pipeline \cite{park2019specaugment}. First we transfer textual user utterance $x$ to its audio form $a$ using gTTS\footnote{\url{https://pypi.org/project/gTTS/}} \cite{oord2016wavenet}, a Text-to-Speech system. Then audio data is translated back into text $x'$ by DeepSpeech2 \cite{amodei2016deep}, an Automatic Speech Recognition (ASR) system. We directly use the released models in the DeepSpeech2 repository\footnote{\url{https://github.com/PaddlePaddle/DeepSpeech}} with the original configuration, where the speech model is trained on Baidu Internal English Dataset, and the language model is trained on CommonCrawl Data.

\begin{table}[!thb]
    \small
    \begin{tabular}{c|c|c}
    \hline
    Type & Original & Augmented \\
    \hline
    Similar sounds & leicester & lester\\
    Liaison & for 3 people & free people \\
    Spoken numbers & 13:45 & thirteen forty five\\
    \hline
    \end{tabular}
    \caption{Examples of speech recognition perturbation.}
    \label{tab:SR_types}
\end{table}

Table \ref{tab:SR_types} shows some typical examples of our SR augmentation. ASR sometimes wrongly identifies one word as another with similar pronunciation. Liaison constantly occurs between successive words. Expressions with numbers including time and price are written in numerical form but different in spoken language.

Since SR may modify the slot values in the translated utterances, fuzzy value detection is employed here to handle similar sounds and liaison problems when it extracts slot values to obtain a semantic label $y'$. However, we do not replace the noisy value with the original value as we encourage such misrecognition in SR, thus $y' \neq y$ is allowed. Moreover, numerical terms are normalized to deal with the spoken number problem. Most slot values could be relocated by our automatic value detection rules. The remainder slot values which vary too much to recognize are discarded along with their corresponding labels.

\paragraph{Speech Disfluency}\label{para:SD}
Disfluency is a common feature of spoken language. We follow the categorization of disfluency in previous works \cite{lickley1995missing,wang2020multi}: filled pauses, repeats, restarts, and repairs. 

\begin{table}[h!tb]
    \small
    \begin{tabular}{@{~}c@{~}|@{~}p{6.2cm}@{~}}
    \hline
    Original & I want to go to Cambridge. \\
    \hline
    Pauses & I want to \textcolor{red}{um} go to \textcolor{red}{uh} Cambridge.\\
    Repeats & I\textcolor{red}{, I} want to go to\textcolor{red}{, go to}  Cambridge.\\
    Restarts & \textcolor{red}{I just} I want to go to Cambridge. \\
    Repairs & I want to go to \textcolor{red}{Liverpool, sorry I mean} Cambridge.  \\
    \hline
    \end{tabular}
    \caption{Example of four types of speech disfluency.}
    \label{tab:SD_types}
\end{table}

We present some examples of SD in Table \ref{tab:SD_types}.
Filler words (``um'', ``uh'') are injected into the sentence to present pauses. 
Repeats are inserted by repeating the previous word. 
In order to approximate the real distribution of disfluency, the \textit{interruption points} of filled pauses and repeats are predicted by a Bi-LSTM+CRF model \cite{zayats2016disfluency} trained on an annotated dataset SwitchBoard \cite{godfrey1992switchboard}, which was collected from real human talks.
For restarts, we insert \textit{false start terms} (``I just'') as a prefix of the utterance to simulate self-correction. 
In LU task, we apply repairs on slot values to fool the models to predict wrong labels. We take the original slot value as \textit{Repair} (``Cambridge'') and take another value with the same slot name as \textit{Reparandum} (``Liverpool''). 
An \textit{edit term} (``sorry, I mean'') is inserted between \textit{Repair} and \textit{Reparandum} to construct a correction. The filler words, restart terms, and edit terms and their occurrence frequency are all sampled from their distribution in SwitchBoard.

In order to keep the spans of slot values intact, each span is regarded as one whole word. No insertions are allowed to operate inside the span. 
Therefore, SD augmentation do not change the original semantic and labels of the utterance, i.e. $y'=y$.

\section{Experimental Setup}
\label{sec:setup}

\begin{figure*}[!thb]
    \centering
    \begin{subfigure}{\linewidth}\centering
    \includegraphics[width=\linewidth]{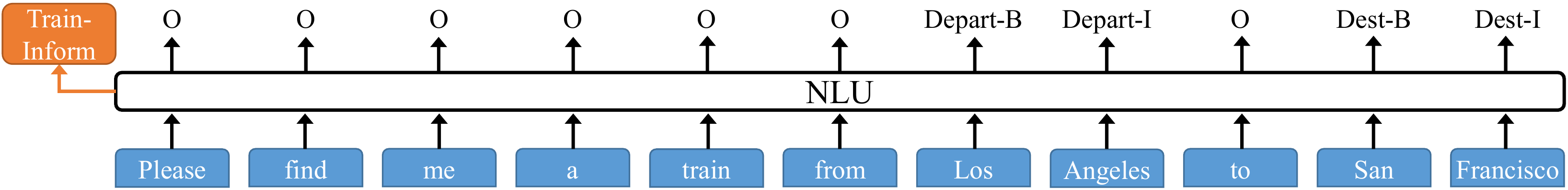}
    \caption{Classification-based language understanding}\label{fig:cls}
    \end{subfigure}
    \begin{subfigure}{\linewidth}\centering
    \includegraphics[width=0.75\linewidth]{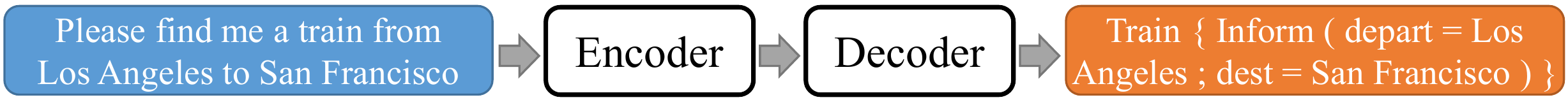}
    \caption{Generation-based language understanding}\label{fig:gen}
    \end{subfigure}
    \caption{An illustration of two categories of language understanding models. Dialog history is first encoded as conditions (not depicted here).}
    \label{fig:nlu}
\end{figure*}

\subsection{Data Preparation}
In our experiments we adopt Frames\footnote{As data division was not defined in Frames, we split the data into training/validation/test set with a ratio of 8:1:1.} \cite{el2017frames} and MultiWOZ \cite{budzianowski2018multiwoz}, which are two task-oriented dialog datasets where semantic labels of user utterances are annotated. In particular, MultiWOZ is one of the most challenging datasets due to its multi-domain setting and complex ontology, and we conduct our experiments on the latest annotation-enhanced version MultiWOZ 2.3 \cite{han2020multiwoz}, which provides cleaned annotations of user dialog acts (i.e. semantic labels). 
The dialog act consists of four parts: domain, intent, slot names, and slot values. The statistics of two datasets are shown in Table \ref{tab:data_statistics}.
Following \citet{takanobu2020your}, we calculate overall F1 scores as evaluation metrics due to the multi-intent setting in LU.
\begin{table}[!thb]
    \small
    \centering
    \begin{tabular}{l|cc}
    \hline
    Datasets & Frames & MultiWOZ \\ 
    \hline
    \# Training Dialogs & 1,095 & 8,438\\
    \# Validation / Test Dialogs & 137 / 137 & 1,000 / 1,000\\
    \# Domains / \# Intents  & 2 / 12 & 7 / 5\\
    Avg. \# Turns per Dialog & 7.60 & 6.85 \\
    Avg. \# Tokens per Turn & 11.67 & 13.55 \\
    Avg. \# DAs per Turn & 1.87 & 1.66 \\
    \hline
    \end{tabular}
    \caption{Statistics of Frames and MultiWOZ 2.3. Only user turns $\mathcal{U}$ are counted here.} 
    \label{tab:data_statistics}
\end{table}

The data are augmented with the inclusion of its copies, leading to a composite of all 4 augmentation types with equal proportion. Other setups are described in each experiment\footnote{See appendix for the hyperparameter setting of LAUG.}.

\begin{table}[h!tb]
    \centering
    \small
    \begin{tabular}{c|ccc|cc}
    \hline
        \multirow{2}{*}{Method}&  
        \multicolumn{3}{c|}{Change Rate/\%}&\multicolumn{2}{c}{Human Annot./\%}\\\cline{2-4}\cline{5-6}
        & Char & Word & Slot &Utter.& DA\\
    \hline
        WP &17.9 &16.0 &36.3 &95.2 &97.0\\
        TP &60.3 &74.4 &13.3 &97.1 &97.7\\
        SR &7.9 &14.5 &40.8 &95.1 &96.7 \\
        SD &22.7 &30.4 &0.4 &98.8 &99.2 \\
    \hline
    \end{tabular}
    \caption{Statistics of augmented MultiWOZ data and their results of quality annotation. Automatic metrics include change rate of characters, words and slot values. Quality evaluation includes appropriateness at utterance level (Utter.) and at dialog act level (DA). }
    \label{tab:aug_stat}
\end{table}

Table \ref{tab:aug_stat} shows the change rates in different aspects by comparing our augmented utterances with the original counterparts. We could find each augmentation method has a distinct effect on the data. For instance, TP rewrites the text without changing the original meaning, thus lexical and syntactic representations dramatically change, while most slot values remain unchanged. In contrast, SR makes the lowest change rate in characters and words but modifies the most slot values due to the speech misrecognition.


\subsection{Quality Evaluation}
To ensure the quality of our augmented test set, we conduct human annotation on 1,000 sampled utterances in each augmented test set of MultiWOZ. We ask annotators to check whether our augmented utterances are reasonable and our auto-detected value annotations are correct (two true-or-false questions). According to the feature of each augmentation method, different evaluation protocols are used. For TP and SD, annotators check whether the meaning of utterances and dialog acts are unchanged. For WP, changing slot values is allowed due to slot value replacement, but the slot name should be the same. For SR, annotators are asked to judge on the similarity of pronunciation rather than semantics. In summary, all the high scores in Table \ref{tab:aug_stat} demonstrate that LAUG makes reasonable augmented examples.

\begin{table*}[ht]
    \small
    \centering
    \begin{subtable}{\linewidth}\centering
    \begin{tabular}{c|c|c|cccc|c|cc}
    \hline
        Model &Train & Ori. & WP & TP & SR & SD & Avg.& Drop & Recov. \\
    \hline
        \multirow{2}{*}{MILU} & Original& 74.15 & 71.05 & 69.58 & 61.53 & 65.27 & 66.86 & -7.29 & /\\ 
        & Augmented& 75.78 & 72.49 & 71.96 & 64.76 & 70.92 & 70.03 & -5.75 & +3.17 \\
        \hline
        \multirow{2}{*}{BERT} & Original& 78.82 & 75.92 & 74.57 & 70.31 & 70.31 & 72.78 & -6.04 & /\\ 
        & Augmented& 78.21 & 76.70 & 75.63 & 72.04 & 77.34 & 75.43 & -2.78 & +2.65  \\
        \hline
        \multirow{2}{*}{ToD-BERT} & Original& \textbf{80.61} & 77.30 & 76.19 & 70.88 & 71.94 & 74.08 & -6.53 & /\\
        & Augmented& 80.37 & \textbf{77.32} & \textbf{77.26} & \textbf{72.54} & \textbf{79.04} & \textbf{76.54} & -3.83 & +2.46 \\
        \hline
        \multirow{2}{*}{CopyNet} & Original& 67.84 & 63.90 & 61.41 & 56.11 & 59.26  & 60.17 & -7.67 & /\\ 
        & Augmented& 69.35 & 67.10 & 65.90 & 60.98 & 67.71  & 65.42 & -3.93 & \textbf{+5.25} \\
        \hline
        \multirow{2}{*}{GPT-2} & Original& 78.78 & 74.96 & 72.85 & 69.00 & 69.19  & 71.50 & -7.28 & /\\ 
        & Augmented& 79.15 & 75.25 & 73.86 & 71.37 & 74.19 & 73.67 & -5.48 & +2.17\\
        \hline
    \end{tabular}
    \caption{Frames}
    \label{tab:frames}
    \end{subtable}
    \begin{subtable}{\linewidth}\centering
    \begin{tabular}{c|c|c|cccc|c|cc}
    \hline
        Model &Train & Ori. & WP & TP & SR & SD & Avg.& Drop & Recov. \\
    \hline
        \multirow{2}{*}{MILU} & Original& 91.33 &  88.26 & 87.20 & 77.98 & 83.67  & 84.28 & -7.05 & /\\
        & Augmented& 91.39 & 90.01 & 88.04 & 86.97 & 89.54 & 88.64 & -2.75 & +4.36 \\
        \hline
        \multirow{2}{*}{BERT} & Original & \textbf{93.40} & 90.96 & 88.51 & 82.35 & 85.98 & 86.95 & -6.45 &/\\
        & Augmented&  93.32 & 92.23 & 89.45 & 89.86 & 92.71 & 91.06 & -2.26 & +4.11 \\
        \hline
        \multirow{2}{*}{ToD-BERT} & Original & 93.28 & 91.27 & 88.95 & 81.16 & 87.18 & 87.14 & -6.14 &/\\
        & Augmented& 93.29 & \textbf{92.40} & 89.71 & \textbf{90.06} & \textbf{92.85} & \textbf{91.26} & -2.03 & +4.12 \\
        \hline
        \multirow{2}{*}{CopyNet} & Original & 90.97 & 85.25 & 87.40 & 71.06 & 77.66 & 80.34 & -10.63 &/\\
        & Augmented& 90.49 & 89.19 & 89.53 & 85.69 & 89.83 & 88.56 & -1.93 &\textbf{+8.22}\\
        \hline
        \multirow{2}{*}{GPT-2} & Original & 91.53 & 85.35 & 88.23 & 80.74 & 84.33 & 84.66 & -6.87 &/ \\
        & Augmented& 91.59 & 90.26 & \textbf{89.92} & 86.55 & 90.55 & 89.32 & -2.27 & +4.66\\
        \hline
    \end{tabular}
    \caption{MultiWOZ}
    \label{tab:multiwoz}
    \end{subtable}
    \caption{Robustness test results. Ori. stands for the original test set, WP, TP, SR, SD for 4 augmented test sets and Avg. for the average performance on 4 augmented test sets. The additional data in augmented training set has the same utterance amount as the original training set and is composed of 4 types of augmented data with equal proportion. Drop shows the performance decline between Avg. and Ori. while Recov. denotes the performance recovery of Avg. between training on augmented/original data (e.g., 88.64\%-84.28\% for MILU on MultiWOZ).}
    \label{tab:main}
\end{table*}

\subsection{Baselines}\label{sec:baseline}

LU models roughly fall into two categories: classification-based and generation-based models.
Classification based models \cite{hakkani2016multi,goo2018slot} extract semantics by intent detection and slot tagging. Intent detection is commonly regarded as a multi-label classification task, and slot tagging is often treated as a sequence labeling task with \textit{BIO format} \cite{ramshaw1999text}, as shown in Fig. \ref{fig:cls}. Generation-based models \cite{liu2016attention,zhao2018improving} generate a dialog act containing intent and slot values. They treat LU as a sequence-to-sequence 
problem and transform a dialog act into a sequential structure as shown in Fig. \ref{fig:gen}.
Five base models with different categories are used in the experiments, as shown in Table \ref{tab:base_model}.

\begin{table}[h!tb]
    \centering
    \small
    \begin{tabular}{l@{}|ccc}
    \hline
        Model & Cls. & Gen. & PLM \\
    \hline
        MILU \cite{hakkani2016multi} & $\surd$ & &\\
        BERT \cite{devlin2019bert} & $\surd$ & & $\surd$ \\
        ToD-BERT \cite{wu2020tod} & $\surd$ &  & $\surd$ \\
        CopyNet \cite{gu2016incorporating} &  & $\surd$ & \\
        GPT-2 \cite{radford2019language} &  & $\surd$ & $\surd$\\
    \hline
    \end{tabular}
    \caption{Features of base models. Cls./Gen. denotes classification/generation-based models. PLM stands for pre-trained language models.}
    \label{tab:base_model}
\end{table}

To support a multi-intent setting in classification-based models, we decouple the LU process as follows: first perform domain classification and intent detection, then concatenate two special tokens which indicate the detected domain and intent (e.g. $[restaurant] [inform]$) at the beginning of the input sequence, and last encode the new sequence to predict slot tags. In this way, the model can address \textit{overlapping slot values} when values are shared in different dialog acts.

\begin{figure*}[h!tb]
    \centering
    \begin{subfigure}{0.5\linewidth}\centering
    \includegraphics[width=\linewidth]{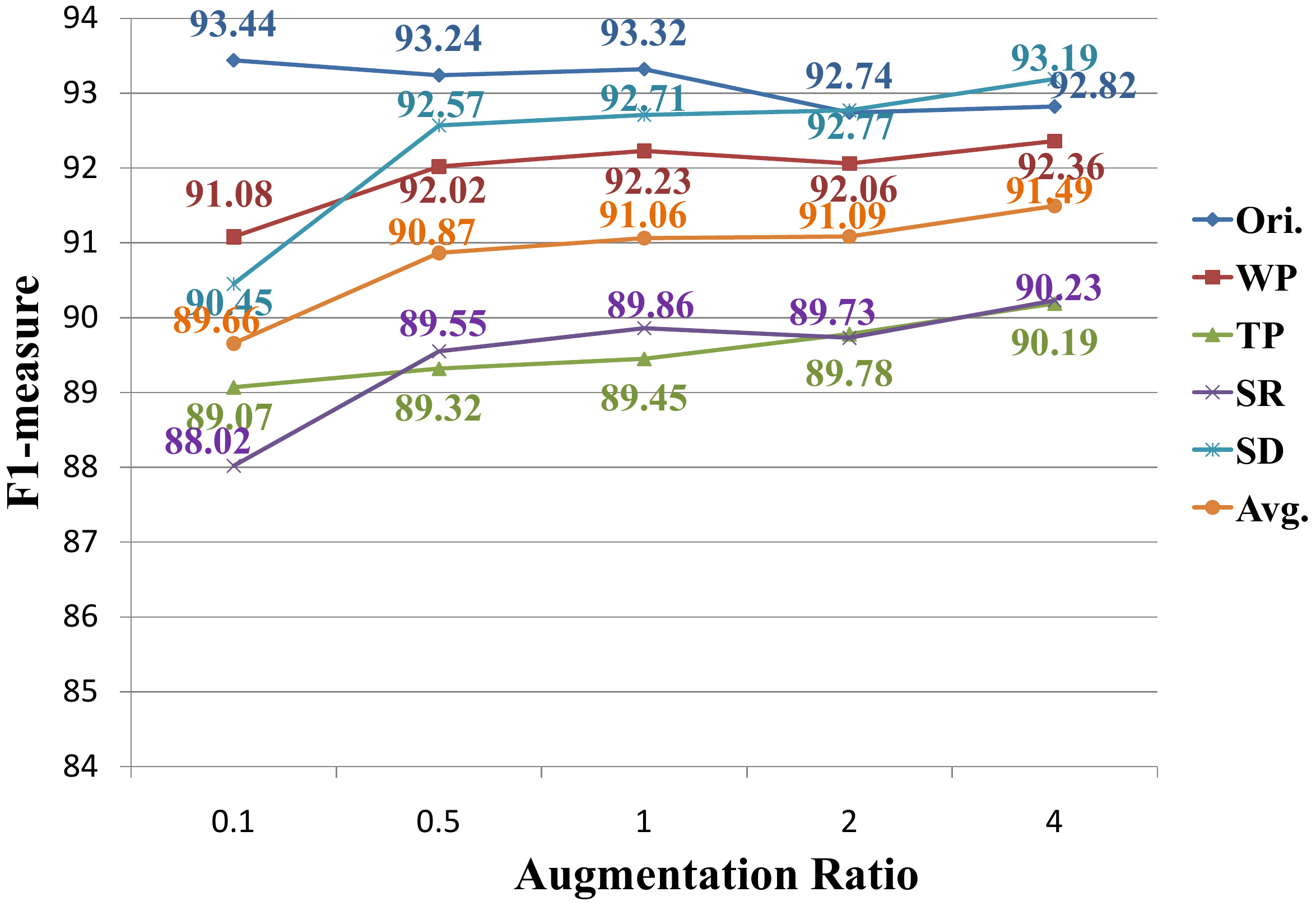}
    \caption{BERT}
    \end{subfigure}%
    \begin{subfigure}{0.5\linewidth}\centering
    \includegraphics[width=\linewidth]{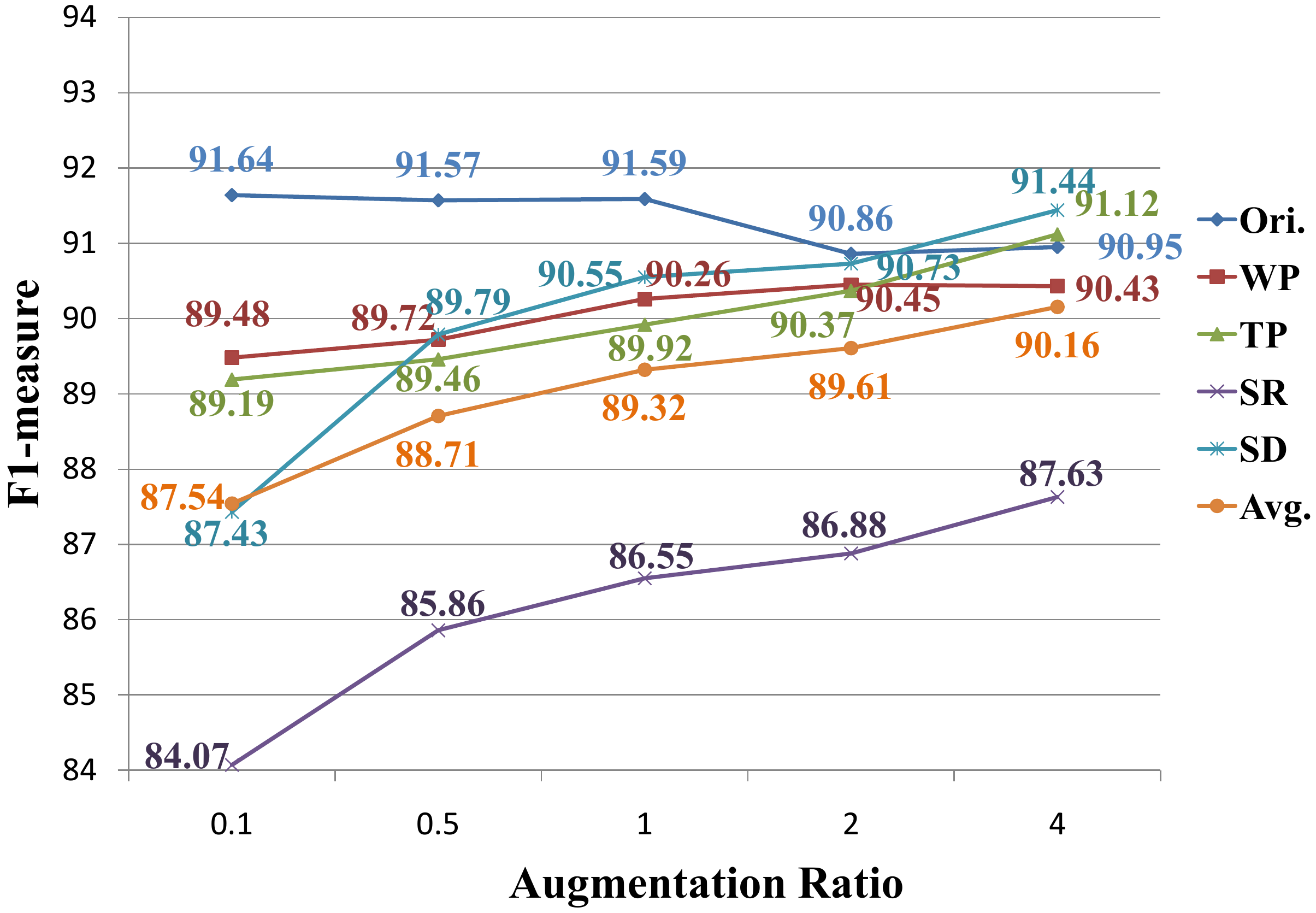}
    \caption{GPT-2}
    \end{subfigure}
    \caption{Performance on MultiWOZ with different ratios of augmented training data amount to the original one. The total amount of training data varies but they are always composed of 4 types of augmented data with even proportion. Different test sets are shown with different colored lines.
    }
    \label{fig:augment}
\end{figure*}


\section{Evaluation Results}\label{sec:result}

\subsection{Main Results}

We conduct robustness testing on all three capacities for five base models using four augmentation methods in LAUG. All baselines are first trained on the original datasets, then finetuned on the augmented datasets. Overall F1-measure performance on Frames and MultiWOZ is shown in Table \ref{tab:main}. All experiments are conducted over 5 runs, and averaged results are reported. 

Robustness for each capacity can be measured by performance drops on the corresponding augmented test sets. All models achieve some performance recovery on augmented test sets after trained on the augmented data, while keeping a comparable result on the original test set. This indicates the effectiveness of LAUG in improving the model's robustness.


We observe that pre-trained models outperform non-pre-trained ones on both original and augmented test sets. 
Classification-based models have better performance and are more robust than generation-based models. 
ToD-BERT, the state-of-the-art model which was further pre-trained on task-oriented dialog data, has comparable performance with BERT. With most augmentation methods, ToD-BERT shows slightly better robustness than BERT.

Since the data volume of Frames is far less than that of MultiWOZ, the performance improvement of pre-trained models on Frames is larger than that on MultiWOZ. Due to the same reason, augmented training data benefits the non-pre-trained models performance of  on Ori. test set more remarkably in Frames where data is not sufficient.

Among the four augmentation methods, SR has the largest impact on the models' performance, and SD comes the second. The dramatic performance drop when testing on SR and SD data indicates that robustness for speech characteristics may be the most challenging issue.


Fig. \ref{fig:augment} shows how the performance of BERT and GPT-2 changes on MultiWOZ when the ratio of augmented training data to the original data varies from 0.1 to 4.0. F1 scores on augmented test sets increase when there are more augmented data for training. The performance of BERT on augmented test sets is improved when augmentation ratio is less than 0.5 but becomes almost unchanged after 0.5 while GPT-2 keeps increasing stably. This result shows the different characteristics between classification-based models and generation-based models when finetuned with augmented data. 

        



\subsection{Ablation Study}

\paragraph{Between augmentation approaches}
In order to study the influence of each augmentation approach in LAUG, we test the performance changes when one augmentation approach is removed from constructing augmented training data. Results on MultiWOZ are shown in Table \ref{tab:between}.

\begin{table}[!th]
    \centering
    \small
    
    \begin{subtable}{\linewidth}\centering
    \begin{tabular}{c|>{\centering}p{0.65cm}|>{\centering}p{0.65cm}>{\centering}p{0.65cm}>{\centering}p{0.65cm}>{\centering}p{0.65cm}|c}
    \hline
        Train & Ori. & WP & TP & SR & SD & Avg. \\
    \hline
        Aug.& 91.39 & 90.01 & 88.04 & 86.97 & 89.54 & 88.64\\
    \hline
        -WP & 91.29 & \textbf{88.42} & 88.43 & 86.98 & 89.20 & 88.26 \\
        -TP & 91.55 & 90.15 & \textbf{87.81} & 86.82 & 89.42 & 88.55\\
        -SR & 91.23 & 90.13 & 88.30 & \textbf{77.90} & 89.51 & 86.46\\
        -SD & 91.56 & 90.24 & 88.60 & 86.78 & \textbf{83.96} & 87.40\\
    \hline
        Ori.& 91.33 &  88.26 & 87.20 & 77.98 & 83.67  & 84.28 \\
    \hline
        
    \end{tabular}
    \caption{MILU}
    \label{tab:between_MILU}
    \end{subtable}
    \begin{subtable}{\linewidth}\centering
    \begin{tabular}{c|>{\centering}p{0.65cm}|>{\centering}p{0.65cm}>{\centering}p{0.65cm}>{\centering}p{0.65cm}>{\centering}p{0.65cm}|c}
    \hline
        Train & Ori. & WP & TP & SR & SD & Avg. \\
    \hline
        Aug. &  93.32 & 92.23 & 89.45 & 89.86 & 92.71 & 91.06 \\
    \hline
        -WP & 93.23 & \textbf{90.94} & 89.42 & 89.93 & 92.82 & 90.78\\
        -TP & 93.08 & 92.24 & \textbf{88.62} & 89.80 & 92.62 & 90.82\\
        -SR & 93.43 & 92.30 & 89.50 & \textbf{83.48} & 93.07 & 89.59\\
        -SD & 93.11 & 92.15 & 89.44 & 90.00 & \textbf{85.22} & 89.20\\
    \hline
        Ori. & 93.40 & 90.96 & 88.51 & 82.35 & 85.98 & 86.95 \\
    \hline
    \end{tabular}
    \caption{BERT}
    \label{tab:between_BERT}
    \end{subtable}
    \caption{Ablation study between augmentation approaches for two models on MultiWOZ. Highlighted numbers denote the most sharp decline for each augmented test set.}
    \label{tab:between}
\end{table}

Large performance decline on each augmented test set is observed when the corresponding augmentation approach is removed in constructing training data. The performance after removing an augmentation method is comparable to the one without augmented training data. Only slight changes are observed without other approaches. These results indicate that our four augmentation approaches are relatively orthogonal. 

\paragraph{Within augmentation approach}
Our implementation of WP and SD consist of several functional components. Ablation experiments here show how much performance is affected by each component in augmented test sets.

\begin{table}[!th]
    \centering
    \small
    \begin{subtable}{\linewidth}\centering
    \begin{tabular}{p{1.05cm}|p{0.6cm}p{0.75cm}|p{0.6cm}p{0.75cm}}
    \hline
       Test & MILU & Diff. & BERT & Diff.\\
    \hline
    
        WP & 88.26 & / & 90.96 & /\\
    \hline
        -Syno. & 88.90 & 0.64 & 91.27 & 0.48 \\
        -Insert & 88.90 & 0.64 & 91.30 & 0.51 \\
        -Delete & 88.97 & 0.71 & 91.20 & 0.41 \\
        -Swap & 89.15 & 0.89 & 91.33 & \textbf{0.54} \\
        -Slot & 89.45 & \textbf{1.19} & 91.30 & 0.51 \\
    \hline
        Ori. & 91.33 & 3.05 & 93.40 & 2.61 \\
    \hline
    \end{tabular} 
    \caption{Word Perturbation}
    \label{tab:eda_ablation}
    \end{subtable}
    \begin{subtable}{\linewidth}\centering
    \begin{tabular}{p{1.13cm}|p{0.6cm}p{0.75cm}|p{0.6cm}p{0.75cm}}
    \hline
        Test & MILU & Diff. & BERT & Diff. \\
    \hline
        SD & 83.67 & / & 85.98 & / \\
    \hline
        -Repair& 89.47 & \textbf{5.80} & 91.05 & \textbf{5.07}\\
        -Pause& 85.21 & 1.54 & 88.06 & 2.08 \\
        -Restart& 84.03 & 0.36 & 86.22 & 0.24 \\
        -Repeat& 83.64 & -0.03 & 85.68 & -0.30 \\
    \hline
        Ori. & 91.33 & 7.66 &93.40 & 7.42\\
    \hline
    \end{tabular}
    \caption{Speech Disfluency}
    \label{tab:disf_ablation}
    \end{subtable}
    \caption{Ablation study within two augmentation approaches. Models are trained on original training set. Highlight stands for the component with the most influence on model performance.}
    \label{tab:within}
\end{table}

Original EDA consists of four functions as described in Table \ref{tab:todeda}. Performance differences (Diff.) can reflect the influences of those components in Table \ref{tab:eda_ablation}. The additional function of our SC-EDA is slot value replacement. We can also observe an increase in performance when it is removed, especially for MILU. This implies a lack of LU robustness in detecting unseen entities.


Table \ref{tab:disf_ablation} shows the results of ablation study on SD. Among the four types of disfluencies described in Table \ref{tab:SD_types}, \textit{repairs} has the largest impact on models' performance. The performance is also affected by \textit{pauses} but to a less extent. The influences of \textit{repeats} and \textit{restarts} are small, which indicates that neural models are robust to handle these two problems.

\subsection{User Evaluation}
In order to test whether the data automatically augmented by LAUG can reflect and alleviate practical robustness problems, we conduct a real user evaluation.
We collected 240 speech utterances from real humans as follows: First, we sampled 120 combinations of DA from the test set of MultiWOZ. Given a combination, each user was asked to speak two utterances with different expressions, in their own language habits.
Then the audio signals were recognized into text using DeepSpeech2, thereby constructing a new test set in real scenarios\footnote{See appendix for details on real data collection.}. Results on this real test set are shown in Table~\ref{tab:user}.

\begin{table}[!htb]
    \centering
    \small
    \begin{tabular}{c|c|ccc}
    \hline
        Model & Train & Ori. & Avg. & Real \\
    \hline
        \multirow{2}{*}{MILU} & Original &91.33 & 84.28 & 63.55 \\
         & Augmented &91.39 & 88.64 & 66.77 \\
         \hline
         \multirow{2}{*}{BERT} & Original &93.40 & 86.95 & 65.22 \\
         & Augmented &93.32 & 91.06 & 69.12 \\
    \hline
    \end{tabular}
    \caption{User evaluation results on MultiWOZ. Ori. and Avg. have the same meaning as the ones in Table \ref{tab:main}, and Real is the real user evaluation set.}
    \label{tab:user}
\end{table}

The performance on the real test set is substantially lower than that on Ori. and Avg., 
indicating that real user evaluation is much more challenging. This is because multiple robustness issues may be included in one real case, while each augmentation method in LAUG evaluates them separately. Despite the difference, model performance on the real data is remarkably improved after every model is finetuned on the augmented data, verifying that LAUG effectively enhances the model's real-world robustness.

\subsection{Error Analysis}

\begin{table}[!thb]
    \centering
    \small
    \begin{tabular}{c|cc|cc}
    \hline
       \multirow{2}{*}{Error Type} & \multicolumn{2}{c|}{BERT Ori.} & \multicolumn{2}{c}{BERT Aug.}\\
       & Num & \% & Num & \% \\
    \hline
       Language Variety &21 & 43.8 &20 & 45.5 \\
       Speech Characteristics &14 & 29.2 &11 & 25.0 \\
       Noise Perturbation & 12& 25.0 &10 & 22.7 \\
       Others &14 & 29.2 &14 & 31.8 \\
    \hline
       Multiple Issues &12 & 25.0 & 11 & 25.0 \\
    \hline
    \end{tabular}
    \caption{Error analysis of BERT in user evaluation.}
    \label{tab:user_error}
\end{table}

Table \ref{tab:user_error} investigates which error type the model has made on the real test set by manually checking all the error outputs of BERT Ori. ``Others'' are the error cases which are not caused by robustness issues, for example, because of the model's poor performance. It can be observed that the model seriously suffers to LU robustness (over 70\%), and that almost half of the error is due to Language Variety. We find that this is because there are more diverse expressions in real user evaluation than in the original data. After augmented training, we can observe that the number of error cases of Speech Characteristics and Noise Perturbation is relatively decreased. This shows that BERT Aug. can solve these two kinds of problems better.
Noting that the sum of four percentages is over 100\% since 25\% error cases involve multiple robustness issues. This again demonstrates that real user evaluation is more challenging than the original test set\footnote{See appendix for case study.}.

\section{Related Work}
\label{sec:prior}


Robustness in LU has always been a challenge in task-oriented dialog. 
Several studies have investigated the model's sensitivity to the collected data distribution, in order to prevent models from overfitting to the training data and improve robustness in the real world.
\citet{kang2018data} collected dialogs with templates and paraphrased with crowd-sourcing to achieve high coverage and diversity in training data. 
\citet{dinan2019build} proposed a training schema that involves human in the loop in dialog systems to enhance the model's defense against human attack in an iterative way.
\citet{ganhotra2020effects} injected natural perturbation into the dialog history manually to refine over-controlled data generated through crowd-sourcing.
All these methods require laborious human intervention. This paper aims to provide an automatic way to test the LU robustness in task-oriented dialog.

Various textual adversarial attacks \cite{zhang2020adversarial} have been proposed and received increasing attentions these years to measure the robustness of a victim model. Most attack methods perform white-box attacks \cite{papernot2016crafting,li2019textbugger,ebrahimi2018hotflip} based on the model's internal structure or gradient signals. Even some black-box attack models are not purely ``black-box'', which require the prediction scores (classification probabilities) of the victim model \cite{jin2020bert,ren2019generating,alzantot2018generating}. However, all these methods address random perturbation but do not consider linguistic phenomena to evaluate the real-life generalization of LU models.

While data augmentation can be an efficient method to address data sparsity, it can improve the generalization abilities and measure the model robustness as well \cite{eshghi2017bootstrapping}.
Paraphrasing that rewrites the utterances in dialog has been used to get diverse representation and thus enhancing robustness \cite{ray2018robust,zhao2019data,iyyer2018adversarial}. 
Word-level operations \cite{kolomiyets2011model,li2020textat,wei2019eda} including replacement, insertion, and deletion were also proposed to increase language variety.
Other studies \cite{shah2019robust,xu2014targeted} worked on the out-of-vocabulary problem when facing unseen user expression. Some other research focused on building robust spoken language understanding \cite{zhu2018robust,henderson2012discriminative,huang2019adapting} from audio signals beyond text transcripts. Simulating ASR errors \cite{schatzmann2007error,park2019specaugment,wang2020data} and speaker disfluency \cite{wang2020multi,qader2018disfluency} can be promising solutions to enhance robustness to voice input when only textual data are provided. As most work tackles LU robustness from only one perspective, we present a comprehensive study to reveal three critical issues in this paper, and shed light on a thorough robustness evaluation of LU in dialog systems.

\section{Conclusion and Discussion}
\label{sec:concl}
In this paper, we present a systematic robustness evaluation of language understanding (LU) in task-oriented dialog from three aspects: \textit{language variety}, \textit{speech characteristics}, and \textit{noise perturbation}. Accordingly, we develop four data augmentation methods to approximate these language phenomena. In-depth experiments and analysis are conducted on MultiWOZ and Frames, with both classification- and generation-based LU models. The performance drop of all models on augmented test data indicates that these robustness issues are challenging and critical, while pre-trained models are relatively more robust to LU.
Ablation studies are carried out to show the effect and orthogonality of each augmentation approach. 
We also conduct a real user evaluation and verifies that our augmentation methods can reflect and help alleviate real robustness problems.

Existing and future dialog models can be evaluated in terms of robustness with our toolkit and data, as our augmentation model does not depend on any particular LU models. Moreover, our proposed robustness evaluation scheme is extensible. In addition to the four approaches in LAUG, more methods to evaluate LU robustness can be considered in the future. 

\section*{Acknowledgments}
This work was partly supported by the NSFC projects (Key project with No.~61936010 and regular project with No.~61876096).
This work was also supported by the Guoqiang Institute of Tsinghua University, with Grant No.~2019GQG1 and 2020GQG0005.
We would like to thank colleagues from HUAWEI for their constant support and valuable discussion.

\bibliography{acl}

\begin{thebibliography}{57}
\expandafter\ifx\csname natexlab\endcsname\relax\def\natexlab#1{#1}\fi

\bibitem[{Alzantot et~al.(2018)Alzantot, Sharma, Elgohary, Ho, Srivastava, and
  Chang}]{alzantot2018generating}
Moustafa Alzantot, Yash Sharma, Ahmed Elgohary, Bo-Jhang Ho, Mani Srivastava,
  and Kai-Wei Chang. 2018.
\newblock Generating natural language adversarial examples.
\newblock In \emph{Proceedings of the 2018 Conference on Empirical Methods in
  Natural Language Processing}, pages 2890--2896.

\bibitem[{Amodei et~al.(2016)Amodei, Ananthanarayanan, Anubhai, Bai,
  Battenberg, Case, Casper, Catanzaro, Cheng, Chen et~al.}]{amodei2016deep}
Dario Amodei, Sundaram Ananthanarayanan, Rishita Anubhai, Jingliang Bai, Eric
  Battenberg, Carl Case, Jared Casper, Bryan Catanzaro, Qiang Cheng, Guoliang
  Chen, et~al. 2016.
\newblock Deep speech 2: End-to-end speech recognition in english and mandarin.
\newblock In \emph{International conference on machine learning}, pages
  173--182.

\bibitem[{Bickmore et~al.(2018)Bickmore, Trinh, Olafsson, O'Leary, Asadi,
  Rickles, and Cruz}]{bickmore2018patient}
Timothy~W Bickmore, Ha~Trinh, Stefan Olafsson, Teresa~K O'Leary, Reza Asadi,
  Nathaniel~M Rickles, and Ricardo Cruz. 2018.
\newblock Patient and consumer safety risks when using conversational
  assistants for medical information: an observational study of siri, alexa,
  and google assistant.
\newblock \emph{Journal of medical Internet research}, 20(9):e11510.

\bibitem[{Budzianowski et~al.(2018)Budzianowski, Wen, Tseng, Casanueva, Ultes,
  Ramadan, and Gasic}]{budzianowski2018multiwoz}
Pawe{\l} Budzianowski, Tsung-Hsien Wen, Bo-Hsiang Tseng, I{\~n}igo Casanueva,
  Stefan Ultes, Osman Ramadan, and Milica Gasic. 2018.
\newblock Multiwoz-a large-scale multi-domain wizard-of-oz dataset for
  task-oriented dialogue modelling.
\newblock In \emph{Proceedings of the 2018 Conference on Empirical Methods in
  Natural Language Processing}, pages 5016--5026.

\bibitem[{Devlin et~al.(2019)Devlin, Chang, Lee, and
  Toutanova}]{devlin2019bert}
Jacob Devlin, Ming-Wei Chang, Kenton Lee, and Kristina Toutanova. 2019.
\newblock Bert: Pre-training of deep bidirectional transformers for language
  understanding.
\newblock In \emph{Proceedings of the 2019 Conference of the North American
  Chapter of the Association for Computational Linguistics: Human Language
  Technologies, Volume 1 (Long and Short Papers)}, pages 4171--4186.

\bibitem[{Dinan et~al.(2019)Dinan, Humeau, Chintagunta, and
  Weston}]{dinan2019build}
Emily Dinan, Samuel Humeau, Bharath Chintagunta, and Jason Weston. 2019.
\newblock Build it break it fix it for dialogue safety: Robustness from
  adversarial human attack.
\newblock In \emph{Proceedings of the 2019 Conference on Empirical Methods in
  Natural Language Processing and the 9th International Joint Conference on
  Natural Language Processing (EMNLP-IJCNLP)}, pages 4529--4538.

\bibitem[{Ebrahimi et~al.(2018)Ebrahimi, Rao, Lowd, and
  Dou}]{ebrahimi2018hotflip}
Javid Ebrahimi, Anyi Rao, Daniel Lowd, and Dejing Dou. 2018.
\newblock Hotflip: White-box adversarial examples for text classification.
\newblock In \emph{Proceedings of the 56th Annual Meeting of the Association
  for Computational Linguistics (Volume 2: Short Papers)}, pages 31--36.

\bibitem[{El~Asri et~al.(2017)El~Asri, Schulz, Sarma, Zumer, Harris, Fine,
  Mehrotra, and Suleman}]{el2017frames}
Layla El~Asri, Hannes Schulz, Shikhar~Kr Sarma, Jeremie Zumer, Justin Harris,
  Emery Fine, Rahul Mehrotra, and Kaheer Suleman. 2017.
\newblock Frames: a corpus for adding memory to goal-oriented dialogue systems.
\newblock In \emph{Proceedings of the 18th Annual SIGdial Meeting on Discourse
  and Dialogue}, pages 207--219.

\bibitem[{Eshghi et~al.(2017)Eshghi, Shalyminov, and
  Lemon}]{eshghi2017bootstrapping}
Arash Eshghi, Igor Shalyminov, and Oliver Lemon. 2017.
\newblock Bootstrapping incremental dialogue systems from minimal data: the
  generalisation power of dialogue grammars.
\newblock In \emph{Proceedings of the 2017 Conference on Empirical Methods in
  Natural Language Processing}, pages 2220--2230.

\bibitem[{Ganhotra et~al.(2020)Ganhotra, Moore, Joshi, and
  Wadhawan}]{ganhotra2020effects}
Jatin Ganhotra, Robert~C Moore, Sachindra Joshi, and Kahini Wadhawan. 2020.
\newblock Effects of naturalistic variation in goal-oriented dialog.
\newblock In \emph{Proceedings of the 2020 Conference on Empirical Methods in
  Natural Language Processing: Findings}, pages 4013--4020.

\bibitem[{Gao et~al.(2019)Gao, Galley, and Li}]{gao2019neural}
Jianfeng Gao, Michel Galley, and Lihong Li. 2019.
\newblock Neural approaches to conversational ai.
\newblock \emph{Foundations and Trends{\textregistered} in Information
  Retrieval}, 13(2-3):127--298.

\bibitem[{Godfrey et~al.(1992)Godfrey, Holliman, and
  McDaniel}]{godfrey1992switchboard}
John~J Godfrey, Edward~C Holliman, and Jane McDaniel. 1992.
\newblock Switchboard: Telephone speech corpus for research and development.
\newblock In \emph{Acoustics, Speech, and Signal Processing, IEEE International
  Conference on}, volume~1, pages 517--520. IEEE Computer Society.

\bibitem[{Goo et~al.(2018)Goo, Gao, Hsu, Huo, Chen, Hsu, and
  Chen}]{goo2018slot}
Chih-Wen Goo, Guang Gao, Yun-Kai Hsu, Chih-Li Huo, Tsung-Chieh Chen, Keng-Wei
  Hsu, and Yun-Nung Chen. 2018.
\newblock Slot-gated modeling for joint slot filling and intent prediction.
\newblock In \emph{Proceedings of the 2018 Conference of the North American
  Chapter of the Association for Computational Linguistics: Human Language
  Technologies, Volume 2 (Short Papers)}, pages 753--757.

\bibitem[{Gu et~al.(2016)Gu, Lu, Li, and Li}]{gu2016incorporating}
Jiatao Gu, Zhengdong Lu, Hang Li, and Victor~OK Li. 2016.
\newblock Incorporating copying mechanism in sequence-to-sequence learning.
\newblock In \emph{Proceedings of the 54th Annual Meeting of the Association
  for Computational Linguistics (Volume 1: Long Papers)}, pages 1631--1640.

\bibitem[{Hakkani-T{\"u}r et~al.(2016)Hakkani-T{\"u}r, Tur, Celikyilmaz, Chen,
  Gao, Deng, and Wang}]{hakkani2016multi}
Dilek Hakkani-T{\"u}r, Gokhan Tur, Asli Celikyilmaz, Yun-Nung Chen, Jianfeng
  Gao, Li~Deng, and Ye-Yi Wang. 2016.
\newblock Multi-domain joint semantic frame parsing using bi-directional
  rnn-lstm.
\newblock \emph{Interspeech 2016}, pages 715--719.

\bibitem[{Han et~al.(2020)Han, Liu, Takanobu, Lian, Huang, Peng, and
  Huang}]{han2020multiwoz}
Ting Han, Ximing Liu, Ryuichi Takanobu, Yixin Lian, Chongxuan Huang, Wei Peng,
  and Minlie Huang. 2020.
\newblock Multiwoz 2.3: A multi-domain task-oriented dataset enhanced with
  annotation corrections and co-reference annotation.
\newblock \emph{arXiv preprint arXiv:2010.05594}.

\bibitem[{He et~al.(2020)He, Yan, and Weiran}]{he2020learning}
Keqing He, Yuanmeng Yan, and XU~Weiran. 2020.
\newblock Learning to tag oov tokens by integrating contextual representation
  and background knowledge.
\newblock In \emph{Proceedings of the 58th Annual Meeting of the Association
  for Computational Linguistics}, pages 619--624.

\bibitem[{Henderson et~al.(2012)Henderson, Ga{\v{s}}i{\'c}, Thomson,
  Tsiakoulis, Yu, and Young}]{henderson2012discriminative}
Matthew Henderson, Milica Ga{\v{s}}i{\'c}, Blaise Thomson, Pirros Tsiakoulis,
  Kai Yu, and Steve Young. 2012.
\newblock Discriminative spoken language understanding using word confusion
  networks.
\newblock In \emph{2012 IEEE Spoken Language Technology Workshop (SLT)}, pages
  176--181. IEEE.

\bibitem[{Honal and Schultz(2003)}]{honal2003correction}
Matthias Honal and Tanja Schultz. 2003.
\newblock Correction of disfluencies in spontaneous speech using a
  noisy-channel approach.
\newblock In \emph{Eighth European Conference on Speech Communication and
  Technology}, pages 2781--2784.

\bibitem[{Huang and Chen(2019)}]{huang2019adapting}
Chao-Wei Huang and Yun-Nung Chen. 2019.
\newblock Adapting pretrained transformer to lattices for spoken language
  understanding.
\newblock In \emph{2019 IEEE Automatic Speech Recognition and Understanding
  Workshop (ASRU)}, pages 845--852. IEEE.

\bibitem[{Iyyer et~al.(2018)Iyyer, Wieting, Gimpel, and
  Zettlemoyer}]{iyyer2018adversarial}
Mohit Iyyer, John Wieting, Kevin Gimpel, and Luke Zettlemoyer. 2018.
\newblock Adversarial example generation with syntactically controlled
  paraphrase networks.
\newblock In \emph{Proceedings of the 2018 Conference of the North American
  Chapter of the Association for Computational Linguistics: Human Language
  Technologies, Volume 1 (Long Papers)}, pages 1875--1885.

\bibitem[{Jin et~al.(2020)Jin, Jin, Tianyi~Zhou, and Szolovits}]{jin2020bert}
Di~Jin, Zhijing Jin, Joey Tianyi~Zhou, and Peter Szolovits. 2020.
\newblock Is bert really robust? a strong baseline for natural language attack
  on text classification and entailment.
\newblock In \emph{Proceedings of the AAAI conference on artificial
  intelligence}, volume~34, pages 8018--8025.

\bibitem[{Kang et~al.(2018)Kang, Zhang, Kummerfeld, Tang, and
  Mars}]{kang2018data}
Yiping Kang, Yunqi Zhang, Jonathan~K Kummerfeld, Lingjia Tang, and Jason Mars.
  2018.
\newblock Data collection for dialogue system: A startup perspective.
\newblock In \emph{Proceedings of the 2018 Conference of the North American
  Chapter of the Association for Computational Linguistics: Human Language
  Technologies, Volume 3 (Industry Papers)}, pages 33--40.

\bibitem[{Kolomiyets et~al.(2011)Kolomiyets, Bethard, and
  Moens}]{kolomiyets2011model}
Oleksandr Kolomiyets, Steven Bethard, and Marie-Francine Moens. 2011.
\newblock Model-portability experiments for textual temporal analysis.
\newblock In \emph{Proceedings of the 49th annual meeting of the association
  for computational linguistics: human language technologies}, volume~2, pages
  271--276.

\bibitem[{Lee et~al.(2019)Lee, Zhu, Takanobu, Zhang, Zhang, Li, Li, Peng, Li,
  Huang et~al.}]{lee2019convlab}
Sungjin Lee, Qi~Zhu, Ryuichi Takanobu, Zheng Zhang, Yaoqin Zhang, Xiang Li,
  Jinchao Li, Baolin Peng, Xiujun Li, Minlie Huang, et~al. 2019.
\newblock Convlab: Multi-domain end-to-end dialog system platform.
\newblock In \emph{Proceedings of the 57th Annual Meeting of the Association
  for Computational Linguistics: System Demonstrations}, pages 64--69.

\bibitem[{Li et~al.(2019)Li, Ji, Du, Li, and Wang}]{li2019textbugger}
Jinfeng Li, Shouling Ji, Tianyu Du, Bo~Li, and Ting Wang. 2019.
\newblock Textbugger: Generating adversarial text against real-world
  applications.
\newblock In \emph{26th Annual Network and Distributed System Security
  Symposium}.

\bibitem[{Li and Qiu(2020)}]{li2020textat}
Linyang Li and Xipeng Qiu. 2020.
\newblock Textat: Adversarial training for natural language understanding with
  token-level perturbation.
\newblock \emph{arXiv preprint arXiv:2004.14543}.

\bibitem[{Lickley(1995)}]{lickley1995missing}
Robin~J Lickley. 1995.
\newblock Missing disfluencies.
\newblock In \emph{Proceedings of the international congress of phonetic
  sciences}, volume~4, pages 192--195.

\bibitem[{Liu and Lane(2016)}]{liu2016attention}
Bing Liu and Ian Lane. 2016.
\newblock Attention-based recurrent neural network models for joint intent
  detection and slot filling.
\newblock \emph{Interspeech 2016}, pages 685--689.

\bibitem[{Liu et~al.(2019)Liu, Meng, Zhang, Zhou, Chen, and Xu}]{liu2019cm}
Yijin Liu, Fandong Meng, Jinchao Zhang, Jie Zhou, Yufeng Chen, and Jinan Xu.
  2019.
\newblock Cm-net: A novel collaborative memory network for spoken language
  understanding.
\newblock In \emph{Proceedings of the 2019 Conference on Empirical Methods in
  Natural Language Processing and the 9th International Joint Conference on
  Natural Language Processing (EMNLP-IJCNLP)}, pages 1050--1059.

\bibitem[{Oord et~al.(2016)Oord, Dieleman, Zen, Simonyan, Vinyals, Graves,
  Kalchbrenner, Senior, and Kavukcuoglu}]{oord2016wavenet}
Aaron van~den Oord, Sander Dieleman, Heiga Zen, Karen Simonyan, Oriol Vinyals,
  Alex Graves, Nal Kalchbrenner, Andrew Senior, and Koray Kavukcuoglu. 2016.
\newblock Wavenet: A generative model for raw audio.
\newblock \emph{arXiv preprint arXiv:1609.03499}.

\bibitem[{Papernot et~al.(2016)Papernot, McDaniel, Swami, and
  Harang}]{papernot2016crafting}
Nicolas Papernot, Patrick McDaniel, Ananthram Swami, and Richard Harang. 2016.
\newblock Crafting adversarial input sequences for recurrent neural networks.
\newblock In \emph{MILCOM 2016-2016 IEEE Military Communications Conference},
  pages 49--54. IEEE.

\bibitem[{Park et~al.(2019)Park, Chan, Zhang, Chiu, Zoph, Cubuk, and
  Le}]{park2019specaugment}
Daniel~S Park, William Chan, Yu~Zhang, Chung-Cheng Chiu, Barret Zoph, Ekin~D
  Cubuk, and Quoc~V Le. 2019.
\newblock Specaugment: A simple data augmentation method for automatic speech
  recognition.
\newblock \emph{Interspeech 2019}, pages 2613--2617.

\bibitem[{Peng et~al.(2020)Peng, Zhu, Li, Li, Li, Zeng, and Gao}]{peng2020few}
Baolin Peng, Chenguang Zhu, Chunyuan Li, Xiujun Li, Jinchao Li, Michael Zeng,
  and Jianfeng Gao. 2020.
\newblock Few-shot natural language generation for task-oriented dialog.
\newblock In \emph{Proceedings of the 2020 Conference on Empirical Methods in
  Natural Language Processing: Findings}, pages 172--182.

\bibitem[{Qader et~al.(2018)Qader, Lecorv{\'e}, Lolive, and
  S{\'e}billot}]{qader2018disfluency}
Raheel Qader, Gw{\'e}nol{\'e} Lecorv{\'e}, Damien Lolive, and Pascale
  S{\'e}billot. 2018.
\newblock Disfluency insertion for spontaneous tts: Formalization and proof of
  concept.
\newblock In \emph{International Conference on Statistical Language and Speech
  Processing}, pages 32--44. Springer.

\bibitem[{Radford et~al.(2019)Radford, Wu, Child, Luan, Amodei, and
  Sutskever}]{radford2019language}
Alec Radford, Jeffrey Wu, Rewon Child, David Luan, Dario Amodei, and Ilya
  Sutskever. 2019.
\newblock Language models are unsupervised multitask learners.
\newblock \emph{OpenAI Blog}, 1(8):9.

\bibitem[{Ramshaw and Marcus(1999)}]{ramshaw1999text}
Lance~A Ramshaw and Mitchell~P Marcus. 1999.
\newblock Text chunking using transformation-based learning.
\newblock In \emph{Natural language processing using very large corpora}, pages
  157--176. Springer.

\bibitem[{Ray et~al.(2018)Ray, Shen, and Jin}]{ray2018robust}
Avik Ray, Yilin Shen, and Hongxia Jin. 2018.
\newblock Robust spoken language understanding via paraphrasing.
\newblock \emph{Interspeech 2018}, pages 3454--3458.

\bibitem[{Ren et~al.(2019)Ren, Deng, He, and Che}]{ren2019generating}
Shuhuai Ren, Yihe Deng, Kun He, and Wanxiang Che. 2019.
\newblock Generating natural language adversarial examples through probability
  weighted word saliency.
\newblock In \emph{Proceedings of the 57th annual meeting of the association
  for computational linguistics}, pages 1085--1097.

\bibitem[{Ribeiro et~al.(2020)Ribeiro, Wu, Guestrin, and
  Singh}]{ribeiro2020beyond}
Marco~Tulio Ribeiro, Tongshuang Wu, Carlos Guestrin, and Sameer Singh. 2020.
\newblock Beyond accuracy: Behavioral testing of nlp models with checklist.
\newblock In \emph{Proceedings of the 58th Annual Meeting of the Association
  for Computational Linguistics}, pages 4902--4912.

\bibitem[{Schatzmann et~al.(2007)Schatzmann, Thomson, and
  Young}]{schatzmann2007error}
Jost Schatzmann, Blaise Thomson, and Steve Young. 2007.
\newblock Error simulation for training statistical dialogue systems.
\newblock In \emph{2007 IEEE Workshop on Automatic Speech Recognition \&
  Understanding (ASRU)}, pages 526--531. IEEE.

\bibitem[{Shah et~al.(2019)Shah, Gupta, Fayazi, and
  Hakkani-Tur}]{shah2019robust}
Darsh Shah, Raghav Gupta, Amir Fayazi, and Dilek Hakkani-Tur. 2019.
\newblock Robust zero-shot cross-domain slot filling with example values.
\newblock In \emph{Proceedings of the 57th Annual Meeting of the Association
  for Computational Linguistics}, pages 5484--5490.

\bibitem[{Takanobu et~al.(2020)Takanobu, Zhu, Li, Peng, Gao, and
  Huang}]{takanobu2020your}
Ryuichi Takanobu, Qi~Zhu, Jinchao Li, Baolin Peng, Jianfeng Gao, and Minlie
  Huang. 2020.
\newblock Is your goal-oriented dialog model performing really well? empirical
  analysis of system-wise evaluation.
\newblock In \emph{Proceedings of the 21th Annual Meeting of the Special
  Interest Group on Discourse and Dialogue}, pages 297--310.

\bibitem[{Wang et~al.(2020{\natexlab{a}})Wang, Fazel-Zarandi, Tiwari,
  Matsoukas, and Polymenakos}]{wang2020data}
Longshaokan Wang, Maryam Fazel-Zarandi, Aditya Tiwari, Spyros Matsoukas, and
  Lazaros Polymenakos. 2020{\natexlab{a}}.
\newblock Data augmentation for training dialog models robust to speech
  recognition errors.
\newblock In \emph{Proceedings of the 2nd Workshop on Natural Language
  Processing for Conversational AI}, pages 63--70.

\bibitem[{Wang et~al.(2020{\natexlab{b}})Wang, Che, Liu, Qin, Liu, and
  Wang}]{wang2020multi}
Shaolei Wang, Wangxiang Che, Qi~Liu, Pengda Qin, Ting Liu, and William~Yang
  Wang. 2020{\natexlab{b}}.
\newblock Multi-task self-supervised learning for disfluency detection.
\newblock In \emph{Proceedings of the AAAI Conference on Artificial
  Intelligence}, volume~34, pages 9193--9200.

\bibitem[{Wang et~al.(2018)Wang, Shen, and Jin}]{wang2018bi}
Yu~Wang, Yilin Shen, and Hongxia Jin. 2018.
\newblock A bi-model based rnn semantic frame parsing model for intent
  detection and slot filling.
\newblock In \emph{Proceedings of the 2018 Conference of the North American
  Chapter of the Association for Computational Linguistics: Human Language
  Technologies, Volume 2 (Short Papers)}, pages 309--314.

\bibitem[{Wei and Zou(2019)}]{wei2019eda}
Jason Wei and Kai Zou. 2019.
\newblock Eda: Easy data augmentation techniques for boosting performance on
  text classification tasks.
\newblock In \emph{Proceedings of the 2019 Conference on Empirical Methods in
  Natural Language Processing and the 9th International Joint Conference on
  Natural Language Processing (EMNLP-IJCNLP)}, pages 6383--6389.

\bibitem[{Wu et~al.(2020)Wu, Hoi, Socher, and Xiong}]{wu2020tod}
Chien-Sheng Wu, Steven~CH Hoi, Richard Socher, and Caiming Xiong. 2020.
\newblock Tod-bert: Pre-trained natural language understanding for
  task-oriented dialogue.
\newblock In \emph{Proceedings of the 2020 Conference on Empirical Methods in
  Natural Language Processing (EMNLP)}, pages 917--929.

\bibitem[{Xu and Sarikaya(2014)}]{xu2014targeted}
Puyang Xu and Ruhi Sarikaya. 2014.
\newblock Targeted feature dropout for robust slot filling in natural language
  understanding.
\newblock In \emph{Interspeech 2014}, pages 258--262.

\bibitem[{Yoo et~al.(2019)Yoo, Shin, and Lee}]{yoo2019data}
Kang~Min Yoo, Youhyun Shin, and Sang-goo Lee. 2019.
\newblock Data augmentation for spoken language understanding via joint
  variational generation.
\newblock In \emph{Proceedings of the AAAI conference on artificial
  intelligence}, volume~33, pages 7402--7409.

\bibitem[{Zayats et~al.(2016)Zayats, Ostendorf, and
  Hajishirzi}]{zayats2016disfluency}
Vicky Zayats, Mari Ostendorf, and Hannaneh Hajishirzi. 2016.
\newblock Disfluency detection using a bidirectional lstm.
\newblock \emph{Interspeech 2016}, pages 2523--2527.

\bibitem[{Zhang et~al.(2020{\natexlab{a}})Zhang, Sheng, Alhazmi, and
  Li}]{zhang2020adversarial}
Wei~Emma Zhang, Quan~Z Sheng, Ahoud Alhazmi, and Chenliang Li.
  2020{\natexlab{a}}.
\newblock Adversarial attacks on deep-learning models in natural language
  processing: A survey.
\newblock \emph{ACM Transactions on Intelligent Systems and Technology (TIST)},
  11(3):1--41.

\bibitem[{Zhang et~al.(2020{\natexlab{b}})Zhang, Takanobu, Zhu, Huang, and
  Zhu}]{zhang2020recent}
Zheng Zhang, Ryuichi Takanobu, Qi~Zhu, Minlie Huang, and Xiaoyan Zhu.
  2020{\natexlab{b}}.
\newblock Recent advances and challenges in task-oriented dialog systems.
\newblock \emph{Science China Technological Sciences}, pages 1--17.

\bibitem[{Zhao and Feng(2018)}]{zhao2018improving}
Lin Zhao and Zhe Feng. 2018.
\newblock Improving slot filling in spoken language understanding with joint
  pointer and attention.
\newblock In \emph{Proceedings of the 56th Annual Meeting of the Association
  for Computational Linguistics (Volume 2: Short Papers)}, pages 426--431.

\bibitem[{Zhao et~al.(2019)Zhao, Zhu, and Yu}]{zhao2019data}
Zijian Zhao, Su~Zhu, and Kai Yu. 2019.
\newblock Data augmentation with atomic templates for spoken language
  understanding.
\newblock In \emph{Proceedings of the 2019 Conference on Empirical Methods in
  Natural Language Processing and the 9th International Joint Conference on
  Natural Language Processing (EMNLP-IJCNLP)}, pages 3628--3634.

\bibitem[{Zhu et~al.(2020)Zhu, Zhang, Fang, Li, Takanobu, Li, Peng, Gao, Zhu,
  and Huang}]{zhu2020convlab}
Qi~Zhu, Zheng Zhang, Yan Fang, Xiang Li, Ryuichi Takanobu, Jinchao Li, Baolin
  Peng, Jianfeng Gao, Xiaoyan Zhu, and Minlie Huang. 2020.
\newblock Convlab-2: An open-source toolkit for building, evaluating, and
  diagnosing dialogue systems.
\newblock In \emph{Proceedings of the 58th Annual Meeting of the Association
  for Computational Linguistics: System Demonstrations}, pages 142--149.

\bibitem[{Zhu et~al.(2018)Zhu, Lan, and Yu}]{zhu2018robust}
Su~Zhu, Ouyu Lan, and Kai Yu. 2018.
\newblock Robust spoken language understanding with unsupervised asr-error
  adaptation.
\newblock In \emph{2018 IEEE International Conference on Acoustics, Speech and
  Signal Processing (ICASSP)}, pages 6179--6183. IEEE.

\end{thebibliography}
\bibliographystyle{acl_natbib}

\newpage
\appendix

\section{Experimental Setup}

\subsection{Hyperparameters}
As for hyperparameters in LAUG, we set the ratio of perturbation \textbf{n}umber to text \textbf{l}ength $\alpha=n/l = 0.1$ in EDA . The learning rate used to finetune SC-GPT in TP is 1e-4, the number of training epoch is 5, and the beam size during inference is 5. In SR, the beam size of the language model in DeepSpeech2 is set to 50. The learning rate of Bi-LSTM+CRF in SD is 1e-3. The threshold of fuzzy matching in automatic value detection is set to 0.9 in TP and 0.7 in SR. 

For hyperparameters of base models. The learning rate is set to 1e-4 for BERT, 1e-5 for GPT2, and 1e-3 for MILU and CopyNet. The beam-size of GPT2 and CopyNet is 5 during the decoding step.

\subsection{Real Data Collection}
Among the 120 sampled DA combinations, each combination contains 1 to 3 DAs. Users can organize the DAs in any order provided that they describe DAs with the correct meaning so as to imitate \textbf{diverse} user expressions in real scenarios. Users are also asked to keep \textbf{natural} in both intonation and expression, and communication noise caused by users in speech and language is included during collection. The audios are recorded by users' PCs under their \textbf{real} environmental noises. We use the same settings of DeepSpeech2 in SR to recognize the collected audios. After automatic span detection (also the same as SR's) are applied, we conduct human check and annotation to ensure the quality of labels.

\section{Evaluation Results}

\subsection{Prediction Schemes}

\begin{table}[!th]
    \centering
    \small
    \begin{tabular}{c|c|@{}c@{}|cc|c}
    \hline
        Model & Train & Scheme & Ori. & Avg. & Drop\\
    \hline
        \multirow{4}{*}{MILU} & \multirow{2}{*}{Ori.} & coupled & 85.52 & 82.91 & -2.61\\
        &  & decoupled & \textbf{91.33} & \textbf{84.28} & -7.05\\
    \cline{2-6}
        & \multirow{2}{*}{Aug.} & coupled & 90.00 & 88.15 & -1.85\\
        & & decoupled & \textbf{91.39} & \textbf{88.64} & -2.75\\
    \hline
        \multirow{4}{*}{BERT} & \multirow{2}{*}{Ori.} & coupled & 88.94 & 80.33 & -8.61\\
        &  & decoupled & \textbf{93.40} & \textbf{86.95} & -6.45\\
    \cline{2-6}
        & \multirow{2}{*}{Aug.} & coupled & 88.84 & 88.63 & -0.21\\
        & & decoupled & \textbf{93.32} & \textbf{91.06} & -2.26\\
    \hline
    \end{tabular}
    \caption{Robustness on different schemes on MultiWOZ. The coupled scheme predicts dialog acts with a joint tagging scheme; the decoupled scheme first detects domains and intents, then recognizes the slot tags.}
    \label{tab:decouple}
\end{table}

\begin{table*}[!th]
    \small
    \centering
    \begin{tabular}{c|p{14cm}}
    \hline
        Ori.& I 'm leaving from Leicester and should arrive in Cambridge by 13:45.  \\
    \hline\
        Golden & train \{ inform ( dest = cambridge ; arrive = 13:45 ; depart = leicester ) \}\\
    \hline
    \hline
        WP& I 'm leaving from Leicester and \textcolor{red}{\{in\}$_{swap}$} arrive \textcolor{red}{\{should\}$_{swap}$} Cambridge by \textcolor{red}{\{06:54\}$_{replace}$}.  \\
    \hline\
        Golden & train \{ inform ( dest = cambridge ; arrive = 06:54 ; depart = leicester ) \}\\
    \hline
        MILU& train \{ inform ( dest = cambridge ; arrive = 06:54 ; depart = leicester ) \}\\
        BERT& train \{ inform ( dest = cambridge ; arrive = 06:54 ; depart = leicester ) \}\\
        Copy& train \{ inform ( dest = cambridge ; depart = leicester ) \}\\
        GPT-2& train \{ inform ( dest = cambridge ; arrive = 06:54 ; depart = leicester ) \} \\
    \hline 
    \hline
        TP & \textcolor{red}{Departing from Leicester and going to Cambridge. I need to arrive by 13:45.}\\ 
    \hline
        Golden &train \{ inform ( dest = cambridge ; arrive = 13:45 ; depart = leicester ) \} \\
    \hline
        MILU& train \{ inform ( dest = cambridge ; arrive = 13:45 ; depart = leicester ) \} \\
        BERT& train \{ inform ( dest = cambridge ; arrive = 13:45 ; depart = leicester ) \}\\
        Copy& train \{ inform ( arrive = 13:45 ; depart = leicester ) \} \\
        GPT-2& train \{ inform ( dest = cambridge ; arrive = 13:45 ; depart = leicester ) \} \\
    \hline
    \hline
        SR & I'm leaving from \textcolor{red}{\{lester\}$_{similar}$} and should arrive in Cambridge by \textcolor{red}{\{thirteen forty five\}$_{spoken}$}.\\
    \hline
        Golden & train \{ inform ( dest = cambridge ; arrive = thirteen forty five ; depart = lester ) \}\\
    \hline
        MILU& train \{ inform ( dest = cambridge ) \}\\
        BERT& train \{ inform ( dest = cambridge ; arrive = thirteen forty five ; depart = lester ) \}\\
        Copy&train \{ inform ( dest = \textcolor{blue}{cambridge forty} ; depart = lester ) \}\\
        GPT-2& train \{ inform ( dest = cambridge ; arrive = thirteen forty five ) \}\\
    \hline
    \hline
        SD & \textcolor{red}{\{Well, you know,\}$_{restart}$} I 'm leaving from Leicester and should arrive in \textcolor{red}{\{King's College sorry, i mean\}$_{repair}$} Cambridge by 13:45.\\
    \hline
        Golden &train \{ inform ( dest = cambridge ; arrive = 13:45 ; depart = leicester ) \} \\
    \hline
        MILU& train \{ inform ( dest =  \textcolor{blue}{ king} 
        ; arrive = 13:45 ; depart = leicester ) \}\\
        BERT& train \{ inform ( dest = \textcolor{blue}{king 's college} ; arrive = 13:45 ; depart = leicester ) \}\\
        Copy& train \{ inform (  arrive = 13:45 ; depart = leicester ) \}\\
        GPT-2& train \{ inform ( dest = \textcolor{blue}{king 's college} ; arrive = 13:45 ; depart = leicester  ) \}\\
    \hline
    \end{tabular}
    \caption{Augmented examples and corresponding model outputs. All models are trained on the original data only. Wrong values are colored in blue.}
    \label{tab:case}
\end{table*}

\begin{table*}[!th]
    \small
    \centering
    \begin{tabular}{c|l}
    \hline
        Case-1& The train from Cambridge arrives at seventeen o'clock.  \\
    \hline\
        Golden & train \{ inform ( dest = Cambridge ; arrive = seventeen o'clock  ) \}\\
    
    \hline\
        MILU Ori.& train \{ inform ( dest = Cambridge   ) \}\\
        MILU Aug.& train \{ inform ( dest = Cambridge ; arrive = seventeen o'clock  ) \}\\
        BERT Ori.& train \{ inform ( dest = Cambridge ; arrive = seventeen ) \}\\
        BERT Aug.& train \{ inform ( dest = Cambridge ; arrive = seventeen o'clock  ) \} \\
    \hline 
    \hline
        Case-2& A ticket departs from Cambridge and arrives at Bishops Stortford the police.  \\
    \hline\
        Golden & train \{ inform ( depart = Cambridge ; dest= Bishops Stortford ) \}\\
    
    \hline\
        MILU Ori.& train \{ inform ( depart = Cambridge ; dest= Bishops Stortford ; dest= police)  \}\\
        MILU Aug.& train \{ inform ( depart = Cambridge ; dest= Bishops Stortford )  \}\\
        BERT Ori.& train \{ inform ( depart = Cambridge ; dest= Bishops Stortford )  \}\\
        BERT Aug.& train \{ inform ( depart = Cambridge ; dest= Bishops Stortford )  \} \\
    \hline 
    \hline
        Case-3& How much should I pay for the train ticket? \\
    \hline\
        Golden & train \{ request ( ticket = ? ) \}\\
    \hline\
        MILU Ori.& None \\
        MILU Aug.& train \{ request ( ticket = ? ) \}\\
        BERT Ori.& None\\
        BERT Aug.& None \\
    \hline 
    \end{tabular}
    \caption{User evaluation examples and corresponding model outputs. Ori. and Aug. stand for model before/after augmented training.}
    \label{tab:real_case}
\end{table*}

In this section, we study the influence of training/prediction schemes on LU robustness. As described in Sec. \ref{sec:baseline} of the main paper, the process of classification-based LU models is decoupled into two steps to handle multiple labels: one for domain/intent classification and the other for slot tagging. Another strategy is to use the cartesian product of all the components of dialog acts, which yields a joint tagging scheme as presented in ConvLab \cite{lee2019convlab}. To give an intuitive illustration, the slot tag of the token ``Los'' becomes ``Train-Inform-Depart-B'' in the example described in Fig. \ref{fig:nlu} of the main paper. The classification-based models can predict the dialog acts within a single step in this way.

Table \ref{tab:decouple} shows that MILU and BERT gain from the decoupled scheme on the original test set. This indicates that the decoupled scheme decreases the model complexity by decomposing the output space. Interestingly, there is no consistency between two models in terms of robustness. MILU via the coupled scheme behaves more robustly than the decoupled counterpart (-2.61 vs. -7.05), while BERT with the decoupled scheme outperforms its coupled version in robustness (-6.45 vs. -8.61). 
Meanwhile, BERT benefits from the decoupled scheme and still achieves 86.95\% accuracy, but BERT training with the coupled scheme seems more susceptible. In addition, both MILU and BERT recover more performance by the proposed decoupled scheme. All these results demonstrate the superiority of the decoupled scheme in classification-based LU models.

\subsection{Case Study}

In Table \ref{tab:case}, we present some examples of augmented utterances in MultiWOZ. 
In terms of model performance, MILU, BERT and GPT-2 perform well on WP and TP in the example while CopyNet misses some dialog acts. For the SR utterance, only BERT obtains all the correct labels. MILU and Copynet both fail to find the changed value spans ``lester'' and ``thirteen forty five''. Copynet's copy mechanism is fully confused by recognition error and even predicts discontinuous slot values. GPT-2 successfully finds the non-numerical time but misses ``leseter''. In the SD utterance, the repair term fools all the models. Overall, in this example, BERT performs quite well while MILU and CopyNet expose some of their defects in robustness.

Table \ref{tab:real_case} shows some examples from real user evaluation. In case-1, the user says ``seventeen o'clock'' while time is always represented in numeric formats (e.g. ``17:00'') in the dataset, which is a typical Speech Characteristics problem. Case-2 could be regarded as a Speech Characteristics or Noise Perturbation case because ``please'' is wrongly recognized as ``police'' by ASR models. Case-3 is an example of Language Variety, the user expresses the request of getting ticket price in a different way comparing to the dataset. MILU and BERT failed in most of these cases but fixed some error after augmented training.

\end{document}